FRONT MATTER

## Title

Autonomous Functional Locomotion in a Tendon-Driven Limb via Limited Experience.

## Authors

Ali Marjaninejad[1,2], Darío Urbina-Meléndez[1], Brian A. Cohn[4], Francisco J. Valero-Cuevas[1,2,3,4,5,*]

[*]Corresponding author's email: valero@usc.edu

## Affiliations

Departments of Biomedical[1], Electrical[2], and Aerospace & Mechanical Engineering[3], Department of Computer Science[4], and Division of Biokinesiology & Physical Therapy[5]

University of Southern California, Los Angeles, CA, USA.

## Abstract

Robots will become ubiquitously useful only when they can use few attempts to teach themselves to perform different tasks, even with complex bodies and in dynamical environments. Vertebrates, in fact, successfully use trial-and-error to learn multiple tasks in spite of their intricate tendon-driven anatomies. Roboticists find such tendon-driven systems particularly hard to control because they are simultaneously nonlinear, under-determined (many tendon tensions combine to produce few net joint torques), and over-determined (few joint rotations define how many tendons need to be reeled-in/payed-out). We demonstrate—for the first time in simulation and in hardware—how a model-free approach allows few-shot autonomous learning to produce effective locomotion in a 3-tendon/2-joint tendon-driven leg. Initially, an artificial neural network fed by sparsely sampled data collected using motor babbling creates an inverse map from limb kinematics to motor activations, which is analogous to juvenile vertebrates playing during development. Thereafter, iterative reward-driven exploration of candidate motor activations simultaneously refines the inverse map and finds a functional locomotor limit-cycle autonomously. This biologically-inspired algorithm, which we call G2P (General to Particular), enables versatile adaptation of robots to changes in the target task, mechanics of their bodies, and environment. Moreover, this work empowers future studies of few-shot autonomous learning in biological systems, which is the foundation of their enviable functional versatility.

## Summary

Our General-to-Particular algorithm, as in young vertebrates, uses motor babbling to autonomously learn a desired task within a few attempts.

## MAIN TEXT

Today's successful control algorithms for robots often require accurate models of the plant, task, and environment. Without such models, controllers must either imitate prescribed behavior[1–4] or execute numerous iterations in the real world or in simulation (real-time or off-line) to converge on adequate performance[5,6]. In contrast, an approach that is both model-free and only requires limited interactions becomes necessary when models are not available for complex and changing systems/environments, or when exhaustive iterations are not feasible. Ultimately, model-free approaches that learn using limited interactions with the environment (the so-called "few-shot" problem [7]) could imbue robots with the enviable resilience and versatility of animals during locomotion, manipulation, and flight.

Here, we demonstrate how autonomous learning can produce effective locomotion patterns in a tendon-driven limb (Figure 1 and Figure S1) via a model-free few-shot approach. Our approach is biologically-inspired at two levels: first, by using tendons to generate torque on the joints and second, by using motor babbling—as in juvenile vertebrates[8,9]—to learn the general capabilities of the plant followed by refinements that are particular to a task (i.e., General-to-Particular, or G2P). Moreover, we control the limb via tendons to (i) replicate the general problem biological nervous systems face when controlling limbs[10], and (ii) because tendon-driven limbs can offer robots unique advantages in design, placement of actuators, versatility, and performance[11].

Successful control of tendon-driven limbs is a challenging test of learning and control strategies[12]. Roboticists find such anatomies particularly hard to control because they are simultaneously nonlinear, under-determined (many tendon tensions combine to produce few net joint torques), and over-determined (few joint rotations define how many tendons need to be reeled-in/payed-out)[10,13].

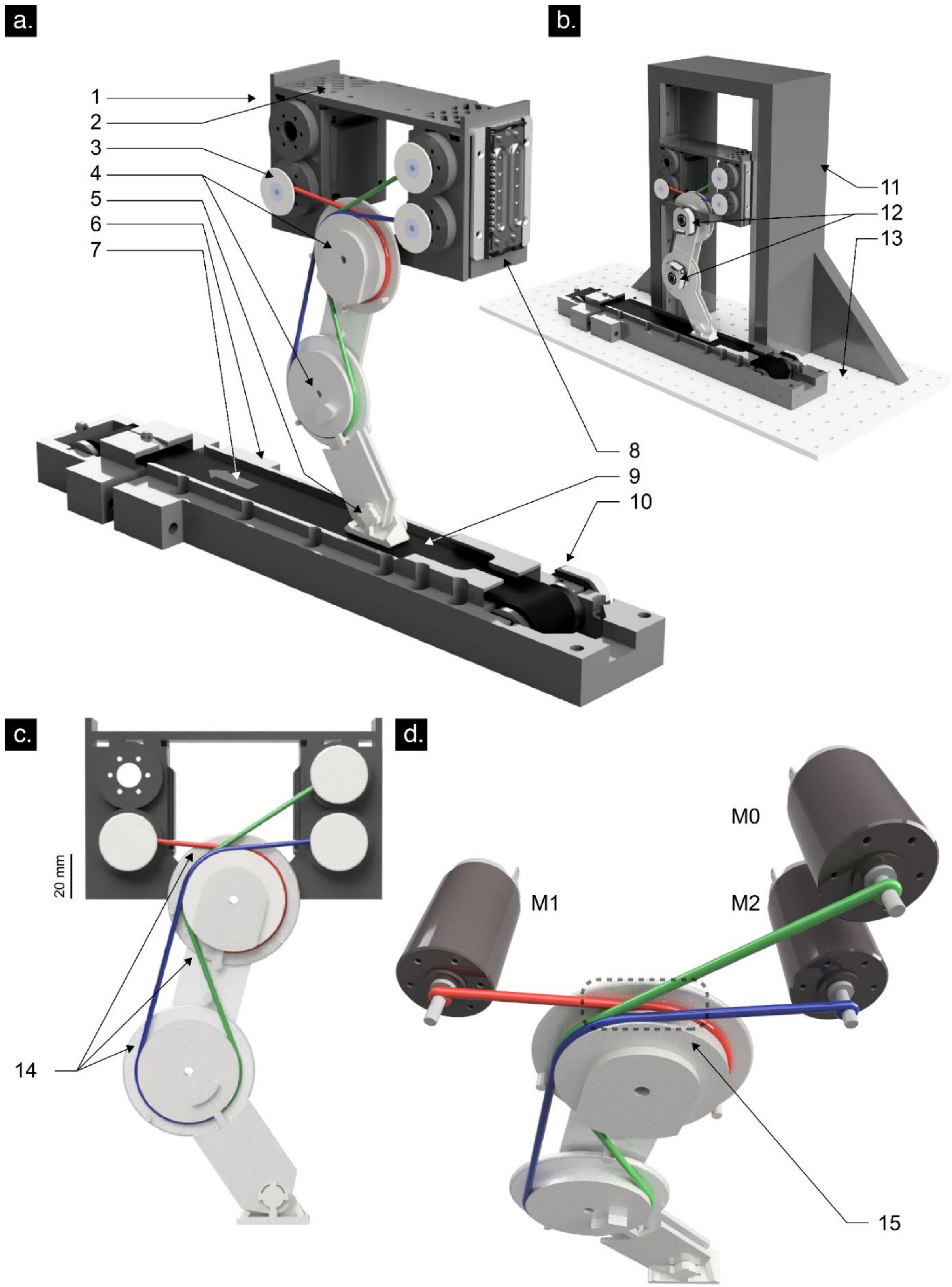

**Figure 1. Planar robotic tendon-driven limb** (a) General system overview 1. Motor-joint carriage 2. Motor ventilation 3. Shaft collars 4. Joints (proximal and distal) 5. Passive hinged foot. 6. Treadmill 7. Direction of positive reward 8. Linear bearings on carriage (locked during testing) 9. Treadmill belt 10. Treadmill drum encoder. (b) Fully supported system 11. Frame 12. Absolute encoders on proximal and distal joints 13. Ground. (c) Tendon routing 14. Three tendons driven by motors M0, M1 and M2. (d) System actuation. Motor M1 drives only the proximal joint ccw, while M0 and M2 drive both joints (M0 drives the proximal joint cw, and the distal joint ccw, while M2 drives both joints cw). 15. Tendon channel.

This work fills a critical gap in robotic learning and control by demonstrating a bio-inspired, practical, and autonomous approach that can perform and adapt to arbitrary tasks, plants and environments, without a priori knowledge using minimal exploration. It also fills a critical gap in computational neuroscience as it provides proof-of-principle of the control of tendon-driven limbs with a biologically and developmentally tenable approach.

## Results

The G2P algorithm autonomously learns locomotion (i.e., propels a treadmill while supported by a carriage) without closed-loop error sensing, nor an explicit model of the dynamics of the tendon-driven limb or ground contact. We also show that execution of multiple attempts can itself lead to improvement in performance on account of a refined inverse map. Such cost-agnostic improvements serve as a proof-of-principle of a biologically-tenable mechanism that benefits from familiarity with the task, rather than teleological optimization, or even error driven correction.

### Locomotion Task

Figure 2 shows an overview of the G2P algorithm described in detail in the Methods. At first, a given run begins with a 5-minute motor babbling session where a time-history of pseudo-random control sequences (a 3-D time-varying vector of current to each motor) is fed to the limb while its kinematics (joint angles, angular velocities and angular accelerations) are measured by encoders at each joint. An artificial neural network then uses these motor babbling data to create an initial inverse map from 6-dimensional kinematics to 3-dimensional control sequences. To produce functional locomotion (without imitation), first, ten free parameters define a limit-cycle feature vector in the joint

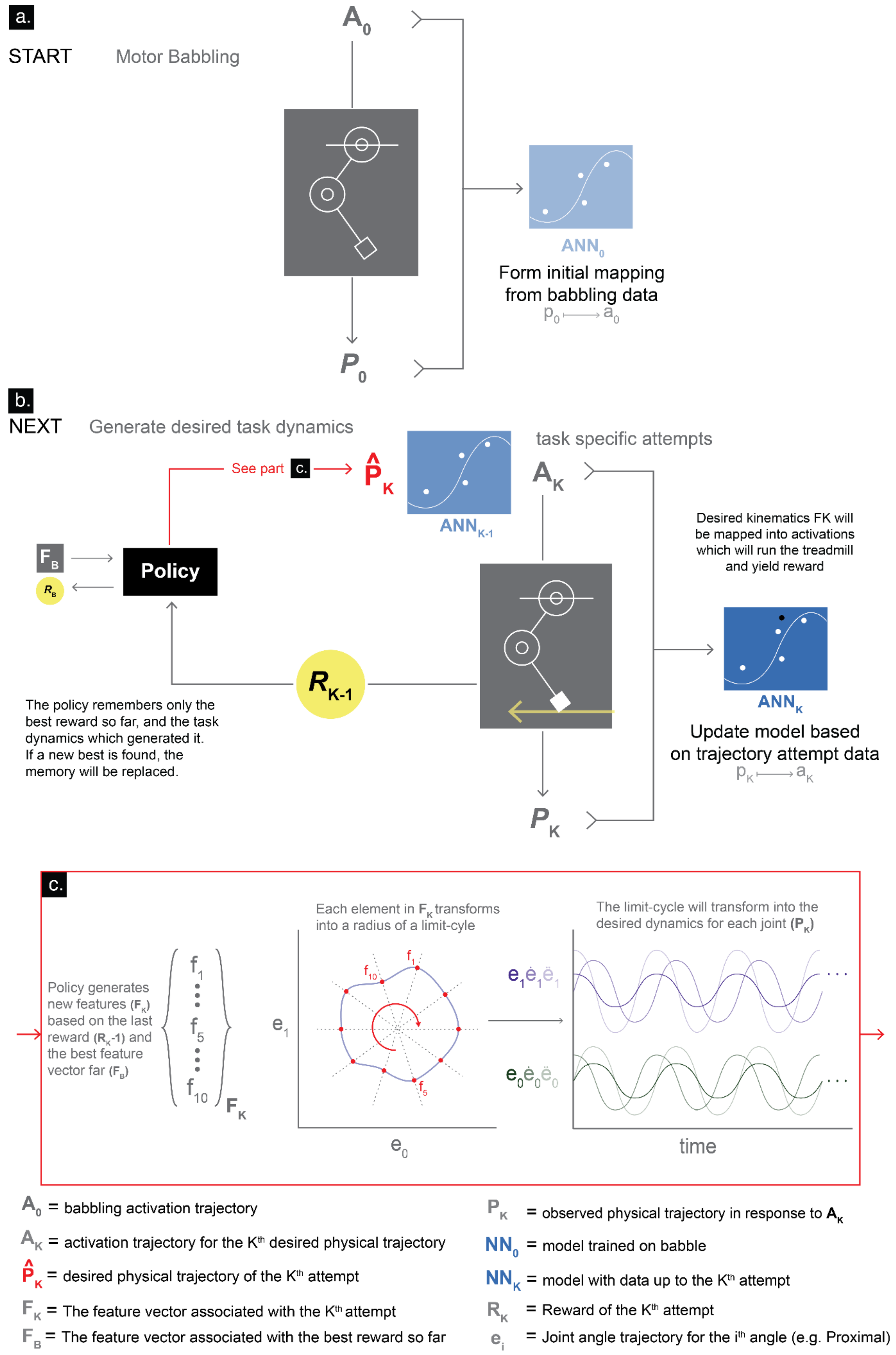

**Figure 2. The G2P algorithm** Every run of the algorithm begins with (a) time-varying babbling control sequences (activations $A_0$ that run through the electric motors) that generate five minutes of random motor

babbling ($P_0$). These input-output data are used to create an inverse (output-input) map $ANN_0$ from limb kinematics to control sequences. (b) Reinforcement learning begins by varying the ten free parameters of the feature vector defining a kinematic limit cycle. These limit cycles in principle can propel the treadmill. $ANN_0$ maps each candidate kinematic limit cycle into activation sequences. An attempt is when a kth activation sequence is repeated twenty times and used to produce twenty steps worth of kinematic data. These kinematic data are further processed and concatenated with all prior data to refine the inverse map into $ANN_K$. The total treadmill propulsion, if any, is the reward for that attempt. (c) If the new reward exceeds the best so far, the policy updates its memory of the best so far and continues its search in the increasingly smaller neighborhood of the most successful limit cycle feature vector. But note that data from all attempts (whether they improve on the best so far or not) are used to refine the inverse map. Figure 3 describes data processing for each run.

angle space (which will define 6-dimensional limb kinematics: joint angles, angular velocities and angular accelerations for each of the two joints; see methods for details). Next, these ten points are interpolated and fed through the initial inverse map which produces a cyclical control sequence that should generate the aforementioned kinematics (Figure 2c). Those predicted control sequences are delivered to the robotic limb twenty times in a row (i.e., for twenty steps or twenty repeats of the locomotor limit cycle). The reward for that attempt is the distance the treadmill was propelled backward, in millimeters (mm), as in forward locomotion. Each run of the G2P algorithm, Figure 3, uses that initial inverse map to start the exploration phase: the ten free parameters of the limit cycle feature vector are changed at random, interpolated, and fed through the inverse map. The resulting control sequences are fed to the motors to produce limb movement until the treadmill reward crosses a threshold of performance set arbitrarily to 64 mm. Thereafter, the exploitation phase of the G2P algorithm, much like a Gaussian Markov process, makes a ten-dimensional probabilistic jump by varying each free parameter from its prior value. If performance improves at a given jump, the standard deviation of those probabilistic jumps is reduced proportionally until it hits a predefined minimum value of 0.03. It is important to note that each time a control sequence is applied (in either the exploration or exploitation phase), the resulting kinematics are recorded, appended to the babbling data and any prior attempts, and a new refined inverse map is calculated (Figure 3b) That is, every interaction with the physics of the plant is used to refine the inverse map. This is analogous to trial-to-trial experiential adaptation during biological motor learning [8].

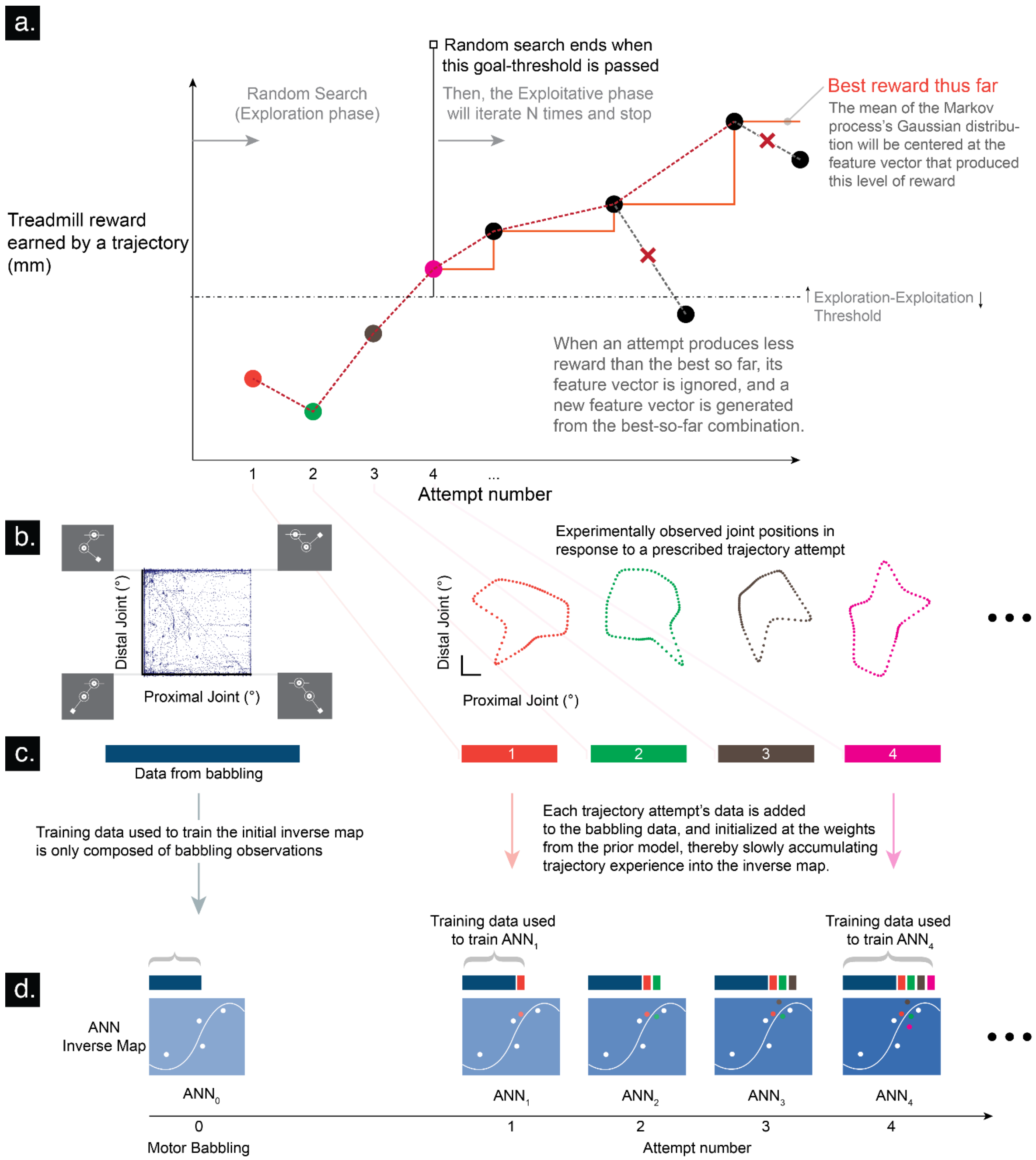

**Figure 3. Reinforcement and Refinement during a given run of G2P** (a) Evolution of reward across the two phases of the reinforcement learning algorithm: the inverse map $ANN_0$ (Figure 2) is used to create the first attempt at a limit cycle defined by the feature vector. The predicted control sequences are applied to the motors to produce twenty cycles of movement that yield a particular treadmill reward (orange dot). The feature vector is modified at random during this exploration phase. This sequence will eventually produce a reward above the exploration-exploitation threshold (dotted line), at which point any changes to the feature vector will be done much like a Markov Gaussian process. (b-c-d) Motor babbling and sequential task-specific refinements of the inverse map: babbling data (enlarged in Figure 6) were used to generate the initial inverse map ($ANN_0$), and are augmented by adding data from each attempt to refine the inverse map after each attempt. (b) distribution of the data in the proximal and distal joints. (c) concatenation of the data and (d) the subsequent refinements of the inverse maps ($ANN_1$, $ANN_2$, …). We limited all runs to have a limit of 15 attempts once the threshold was crossed (see results in Figure 4).

Each point in Figure 4a shows the reward for the corresponding attempt, while a colored stair-step line shows the best reward achieved thus far during that run (fifteen replicates in total, denoted by color). First, our system was able to cross the threshold in a median of 24 attempts in the exploration phase. Second, the attempts in the subsequent exploitation phase showed median reward improvement of 45.5 mm, and with a final median best reward median of 188mm (best run performance was 426.9 mm). Simulation results for the corresponding test are shown on Figure S2.

Figure 4b shows the relationship between the reward received and the energy consumed. Each convex hull contains the family of solutions (i.e., attempts from the exploitation phase which yielded higher than the threshold). Energy expenditure at the end of the exploration phase (triangle) is not systematically different from the energy expenditure of the final solution (large dot). Moreover, high rewards can be achieved with both high and low power consumption. This shows that energy minimization is not an emergent property of this bio-inspired system, and any energy optimization would need to be enforced by a high-level controller.

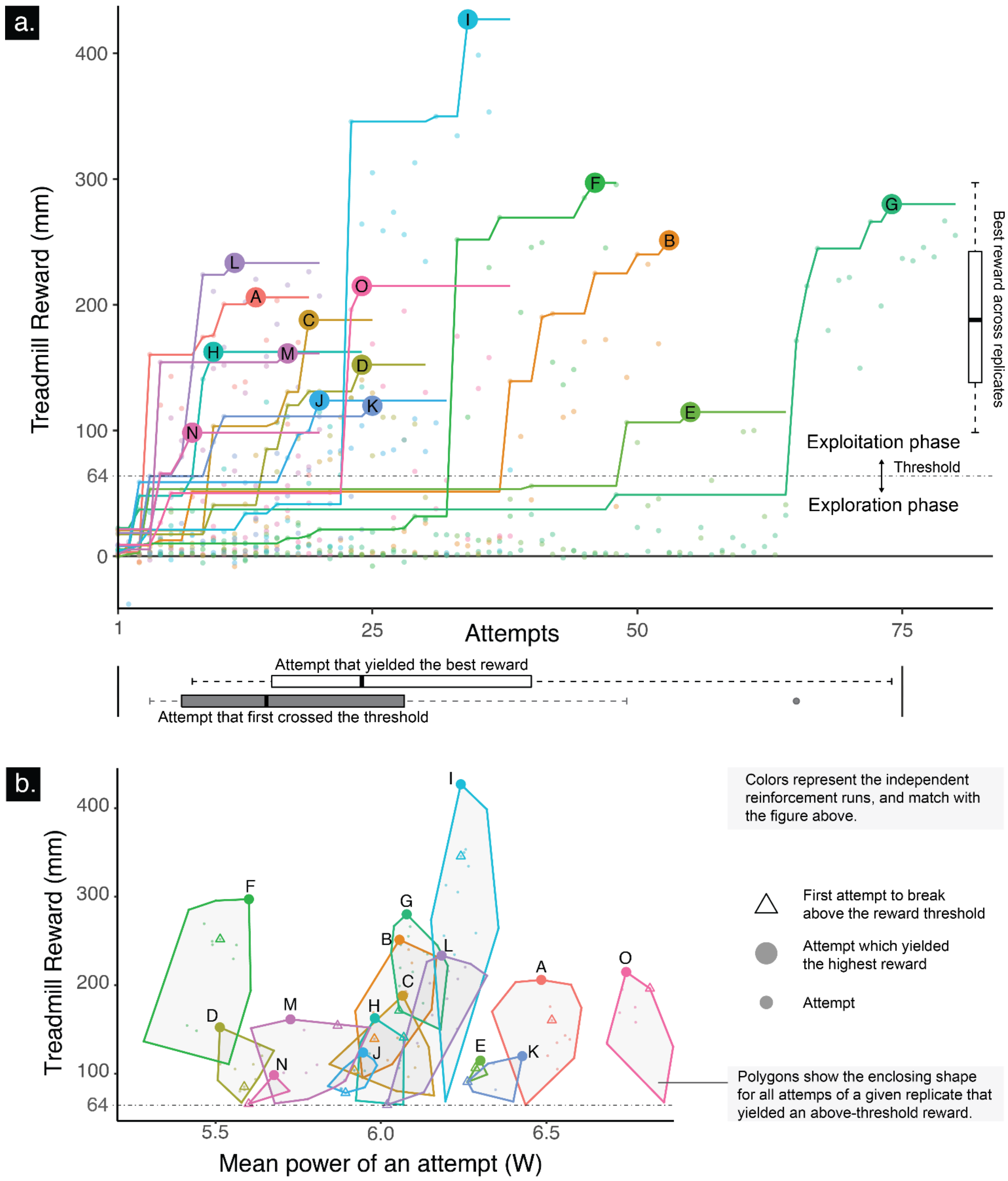

**Figure 4. The treadmill task results** (a) Treadmill reward accrued in fifteen runs, denoted by letters A-O. All runs crossed the exploration-exploitation threshold of 64 mm of treadmill propulsion within a median of fifteen attempts, and the appearance of the best reward happened at a median attempt number of 24. (b) Reward vs. energy consumption, where each polygon includes all attempts in the exploitation phase whose reward was higher than the threshold. We indicate the peak reward within the allowed fifteen exploitation attempts (large dot), as well as the first reward above the exploration-exploitation threshold (triangle).

## Tracking results for free cyclical movements

It is important to note that the utility of the incremental refinements (in increasing the precision of inverse map) cannot be directly interpreted from the results in Figure 4. This is because the reinforcement learning algorithm might, by itself, find a limit-cycle feature vector that yields high reward even with an imprecise inverse map. However, in many applications, such as tracking a desired trajectory (imitation), precision of this inverse map is crucial. In order to evaluate the effect of task specific exploration data in refining the inverse map, we performed the following trajectory tracking tasks.

We used two additional cases to test the ability of the G2P algorithm in tracking desired trajectories (i.e., with no explicit reward).

A. Free cyclical movement in air for a single trajectory

These trials were performed while the leg was suspended 'in the air' to observe trajectory accuracy for an attempt. Treadmill rewards were not collected. As usual, the initial inverse map was extracted from five minutes of motor babbling data, and incrementally refined after each of five attempts performed by the limb (regardless of its tracking error over the course of the attempt). Figure 5 (a-i). shows reduction of the Mean Square Error (MSE) with respect to the limit cycle defined by the feature vector with every attempt. Figures 5 (a-ii,-iii). show the time history of actual achieved vs. desired proximal and distal joint angles for five replicates. Note that Figure 5 (a-ii). is not the endpoint trajectory—it is the joint angle over time as defined by the feature vector of the limit cycle. (see Figure S3a and b for the simulation result of the corresponding test)

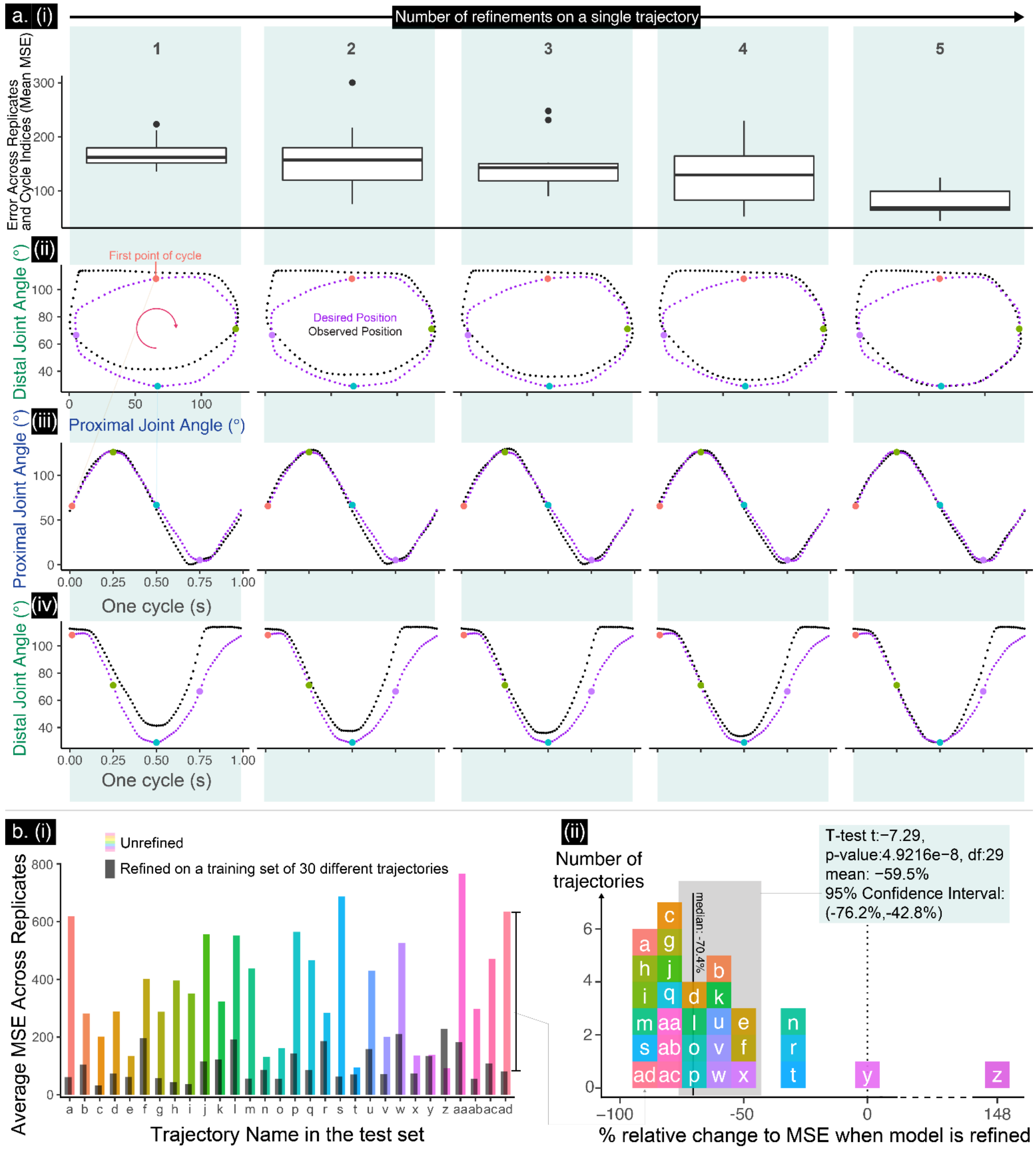

**Figure 5. Tracking results for free cyclical movements** (a) Reduction of MSE during refinements on a fixed trajectory (a-i) Boxplots of the reduction of the Mean Square Error (MSE) with respect to the limit cycle defined by the feature vector with every attempt. (a-ii to a-iv). Desired vs. actual kinematics across refinements on the same fixed trajectory. (b). Test of generalization for first attempt across 30 unseen trajectories a, b, c, …, ad. (see text) (b-i) MSE of the test trajectories using the unrefined (only babble-trained) and the refined (over 30 locomotor trajectories) inverse map. (b-ii) Histogram of percent difference in MSE for the results in (b-i) for each of the 30 unseen trajectories.

B. Generalizability of learned free cyclical movements in air

While the test (A) above explored how repeated refinements can improve the performance of the map in tracking a single desired trajectory, it does not speak to the performance of a given inverse map on unseen trajectories. To evaluate the generalizability of refinement, we followed motor babbling by serial refinement on thirty randomly selected trajectories. This trained inverse map was then "fixed" and evaluated for its MSE accuracy on 30 unseen random trajectories (test set) without further refinement. Figure 5 (b-i,-ii) show that this serially-refined inverse map performed better than when an inverse map trained only by motor babbling was fixed and tested on the same 30 unseen tests. This strongly suggests that refining a map with specific examples does not over-specify its utility, but rather makes it more general and able to produce novel tasks well.

Figures 5(b-i,-ii) show that the inverse map refined on the specific dynamics for a set of free movements generalized to other movements. This is very important since it means the system can learn from every experience and use it in the next attempts which were not explored before. As seen on the figure, the refinement made using 30 random training trajectories improved the MSE in almost all the test trajectories (96.7% of them). This shows that this method is highly generalizable on the trajectories (a-ad) that were not previously explored. Across all test trajectories, the median MSE was 337.18 for the babble-only inverse map, while the median MSE was 83.04 for the refined inverse map; constituting a 75.37% reduction in MSE. (see Figure S3c for the simulation result of the corresponding test)

## Discussion

We introduce the G2P (General to Particular) algorithm by demonstrating how it simultaneously addressed two long-standing problems in robotic control: (i) locomotor function with a tendon-driven limb via (ii) a model-free and few-shot learning approach. The significance of each of these are discussed in detail below. Taken together, our results demonstrate a powerful biologically-inspired approach that can produce versatile adaptation in spite of changes in the particular locomotor trajectory and presence/absence of ground contact. We conclude by outlining how G2P can be combined with computational neuroscience and neuromechanics to understand the biological mechanisms that imbue vertebrate animals with their enviable functional versatility.

### How does G2P move the field forward?

A common approach in robotics is to use models of a system to first develop controllers in simulation (e.g., [14–20]), and then deploy them in physical systems. Creating more accurate models of physical systems is, of course, desirable, but is often impractical since it either requires accurate prior information or extensive physical testing for model inference or system identification. As a result, it is critically important to match the model and simulation environment to the particular application, design and build robots to match available models and capabilities of the simulator[21], or develop robust controllers (which is a field in and of itself)[22,23]. These challenges of model-driven controller design have justified and driven important efforts towards developing model-free or semi-model-free controllers that depend minimally on accurate models, prior information or extensive physical testing. One such approach is model predictive control (MPC). For locomotion and humanoid robots, MPC can provide good performance with limited interactions with the environment in spite of approximate models and unmodeled contact dynamics[5,6]. The current challenges of MPC are that it is difficult to port from simulation to real systems, and it requires extensive computational power, careful design, and sophisticated control theory; furthermore, successful solutions via MPC do not necessarily generalize to other tasks, plants or environments. Such engineering approaches will clearly continue to make headway and improve performance in physical systems[18]. But even as they improve, it is unlikely they can shed light on the biological mechanisms that produce enviable functional versatility in animals such as vertebrates with tendon-driven limbs.

Thus, our work takes an approach to robotic control borne out of the central question in computational neuroscience and neuromechanics[10]: what are the biological mechanisms that have evolved to produce enviable functional versatility within the limitations and complexity of tendon-driven vertebrate limbs?

Fundamentally, vertebrates rely on both brain-body co-evolution (nature) and individual experience (nurture)[24]. At the level of a species, brain-body co-evolution (be it natural or synthetic) adapts (i) the trajectory of structure-function relationships with (ii) its neural circuitry and sensorimotor processing capabilities[25–29]. At the level of a specific individual, initial motor development (during childhood) and experience (throughout the lifespan) continuously enable and refine behavioral capabilities in the context of a changing body or the environment, and for a variety of particular task dynamics and goals—which thereafter takes thousands of repetitions to perfect[9,30]. This first version of the G2P algorithm takes a given brain-body system (i.e., a given species) and focuses on solving how a given individual would learn to make use of its body—even if sub-optimally—within the constraints of limited experience. This is the central question faced by every individual vertebrate at birth: given the particular structure-function relationships lent by morphology, how can it autonomously learn to produce and master the tasks needed for survival? In this study, we challenged the tendon-driven limb to use limited experience to autonomously learn to propel a treadmill (like a leg) and produce a given cyclical movements suspended in air (like an arm or wing).

## Biological inspiration for G2P

The guiding principle in our design of G2P is the fact that individual vertebrates are under strong evolutionary pressure to learn and improve as much as possible from every experience during childhood (development) and thereafter throughout the lifespan to adapt to changes in its body, and new tasks and environments. Vertebrates appear to use processes that hinge on (i) trial-and-error[9,30], memory-based anticipation and kinematic/kinetic pattern recognition (e.g., [31]), (ii) experience-based adaptation (e.g., [32]), and when desired, (iii) using cost-driven reinforcement learning[33]. In a parallel way to how vertebrates learn to use their bodies, the G2P algorithm also uses task-agnostic random inputs to first create a General inverse map of its kinematics, as in (i)—similar to motor babbling and play in young vertebrates (Figure 2). Next, as it begins to perform a task, it refines the inverse map with data coming from every attempt, as in (ii), commanded by the higher-level controller which systematically maximizes a reward by using some form of reinforcement learning, as in (iii). In fact, this continual refinement of a map is reminiscent of synaptic plasticity that creates and reinforces circuits for locomotion in the mammalian spinal cord under the high-level control of the brain[34–36]. This follows the maxim of associative Hebbian learning *"neurons that fire together, wire together"*[37].

More specifically, our results demonstrate that G2P can—in a biologically and developmentally plausible way—simultaneously solve long-standing challenges of controlling physical tendon-driven robots. Namely, G2P can control an inherently nonlinear double pendulum that interacts with the world via intermittent contact, or control interaction torques during free movements. Moreover, since it is controlled by tendons, the system is simultaneously under-determined (many tendon tensions combine to produce few net joint torques) and over-determined (few joint rotations define how many tendons need to be reeled-in/payed-out) as mentioned in the Introduction. We refer the reader to a formal mathematical treatment of the nature of the control of tendon-driven limbs in the context of Feasibility Theory for neuromuscular systems[10,11,38]. Suffice it to say that the control problem the nervous system confronts is to—within a limited number of attempts—find, explore and exploit particular sequences of control actions that inhabit a low-dimensional manifold embedded in high-dimensions. The G2P algorithm is able to confront and solve this same control problem that the nervous system faces.

### Geometric and topological interpretation of G2P

For this robotic limb, the feasible control signals are time-varying 3-D vectors $[u_1(t), u_2(t), u_3(t)]$ of current delivered to the three DC brushless motors (analogous to neural drive to three muscles). Successful control sequences to produce cyclical movements are a one-dimensional limit cycle embedded in 3-D control-signal space that maps into (i.e., produces) a one-dimensional limit cycle embedded in the 6-D space of kinematic variables (i.e., 2 joint angles, angular velocities and angular accelerations of the limb).

A successful control sequence must, by definition: (i) produce time histories of net joint torques (i.e., the underdetermined problem of combining tendon tensions) to satisfy the dynamical equations of motion while (ii) allowing the desired limb kinematics (i.e., the over-determined problem of allowing each motor's axel to reel-in or pay-out tendon lengths as defined by the joint angular velocities). This explains why most motor babbling command signals in the present study (unsurprisingly) produced no limb movement, or rapid movements that run up against the extremes of the range of motion (Figures 3 and 6). The likelihood of finding a control sequence at random that simultaneously satisfies all of these constraints while staying within the interior of the range of motion of the joints is a needle-in-a-haystack problem (i.e., much like throwing a dart that lands on a narrow line in 3-D space[10,11,38], as when infants require hundreds of attempts to learn to transition from sitting to standing[9,30]). Thus, in our robotics limb, most control sequences simply load the tendons against each other and do not produce movement, or abruptly overpower each other and go to the extremes of range of motion (Figures 3 and 6).

Smooth and functional movements will arise only when the G2P algorithm 'implicitly learns' how to solve the over-determined problem of tendon excursions. Such problems have at most one solution—and are usually solved by finding solutions that violate the constraints by "reasonable" (e.g., least-squares) amounts as when calculating a Moore-Penrose pseudoinverse. We have argued that the viscoelastic properties of muscle provide a physical margin of error that is a critical enabler of the neural learning and control of movement[10]. This is precisely why we used backdrivable brushless DC motors: To allow motors to fight such a 'tug-of-war' and find imperfect solutions that can produce movement[39,40]. Had we used servo motors that are stiff, the system would lock-up if tendons are not reeled-in or payed-out precisely. More formally[38,41], the manifold of a 3-dimensional valid control sequence need not be a line (with zero volume), but rather a stretchable tube embedded in 3-D that allows multiple solutions in the neighborhood of an ideal 1-D limit cycle. This is exactly the manifold that G2P is capable of finding through its autonomous and few-shot approach.

Following this reasoning, the many babbling control sequences that 'failed' to produce movement (as 84.3% of the observations in Figure 3a land within 5% of the rotational limit for at least one joint) may nevertheless be informative. One possibility is that such data-driven inverse maps are more than just regressions, and actually provide implicit knowledge of how tendons "fight" each other. Some suggest that data-driven input-output maps are internal representation of invariant properties of the physics of the limb and environment[42,43]. In computational neuroscience, this idea takes the form of an "internal model" used for motor planning and to predict the sensory consequences of motor actions via efference copy[44–47]—which is continually refined by experience. We are agnostic—and make no claims about—whether G2P is learning tendon routings, strain energy[39], or other mechanical properties of the system such as limb impedance[48–50], interaction torques[51], or other dynamical properties[52].

## Familiarity reinforces habits

What could produce improvements in the performance of a task in the absence of real-time feedback or imitation? Every attempt at the movement is used to refine the implicit map, and if successful defines the exploration region, thus we can interpret G2P as analogous to the way in which Hebbian reinforcement tends to cement motor habits. Performance metrics are not used to preferentially weigh data from a given attempt. Rather, familiarity with a task (in the form of all additional data in the neighborhood of the desired movement) reinforces the way it is performed. Motor babbling creates an initial general map, from which a control sequence for a particular movement is extracted. This initial prediction serves as a "belief" about the relationship between body/environment, and an appropriate control strategy. This prediction is used for the first attempt that, while imperfect, does produce additional sensory data now in the neighborhood of the particular task. These data are subsequently leveraged toward refinement of the inverse map, which then leads to an emergent improvement in performance.

Importantly, the details of a given valid solution for movement are idiosyncratic and determined by the first randomly-found control sequence that crossed the exploration-exploitation threshold of performance (Figure 4). Hence all subsequent attempts that produce experience-based refinements are dependent on that seed (much like a Markov process). This solution and its subsequent refinements, in fact, are a family of related solutions can be called a "motor habit" that is adopted and reinforced even though it has no claim to uniqueness nor optimality[53,54]. Biologically speaking, vertebrates also exhibit idiosyncrasies in their motor behavior, which is why it is easy to recognize health states[55], sexual fitness[56], identify individuals by the details of their individual movement and speech habits, and even tell their styles and moods[57,58]. A subtle but important distinction is that these emergent motor habits are not necessarily local minima in the traditional sense. They are good enough solutions that were reinforced by familiarity with a particular way of doing a task. There is evidence that such multiplicity of sub-optimal yet feasible setpoints for the gains in spinal circuitry for discrete and cyclical movements[36]. Those authors argue that is evolutionary advantageous for vertebrates to inherit a body that is easy to learn to control by adopting idiosyncratic, yet useful, motor habits created and reinforced by an individual's own limited experience, without consideration of optimality[54].

To further explore the notion of familiarity as an enabler of learning, we performed experiments testing the ability of the leg to produce free cyclical movements in air, without contact with the treadmill—and hence without explicit reward. That is, we already knew that the motor babbling session could inform the production of a movement, even though 84.3% of babbling has at least one joint “stuck” at limit of its range of motion, Figure 3a. But we can say, as shown in Figure 5, that performance of a particular free cyclical movement improves simply on the basis of repeated attempts. This represents, essentially, the cementing of a motor habit on the basis of experience in the neighborhood of the particular movement.  Figure 6 further shows how 15 cycles of a particular target task in the interior of the joint angle space, which is the most poorly explored region during babbling.  Note that the absence of a reward or penalty allowed the emergent solution to contain a portion where the distal joint is at its limit of range of motion. This, however, need not be detrimental to behavior. For example, human walking can also have the knee locked in full extension during the latter part of the swing phase right before heel strike.

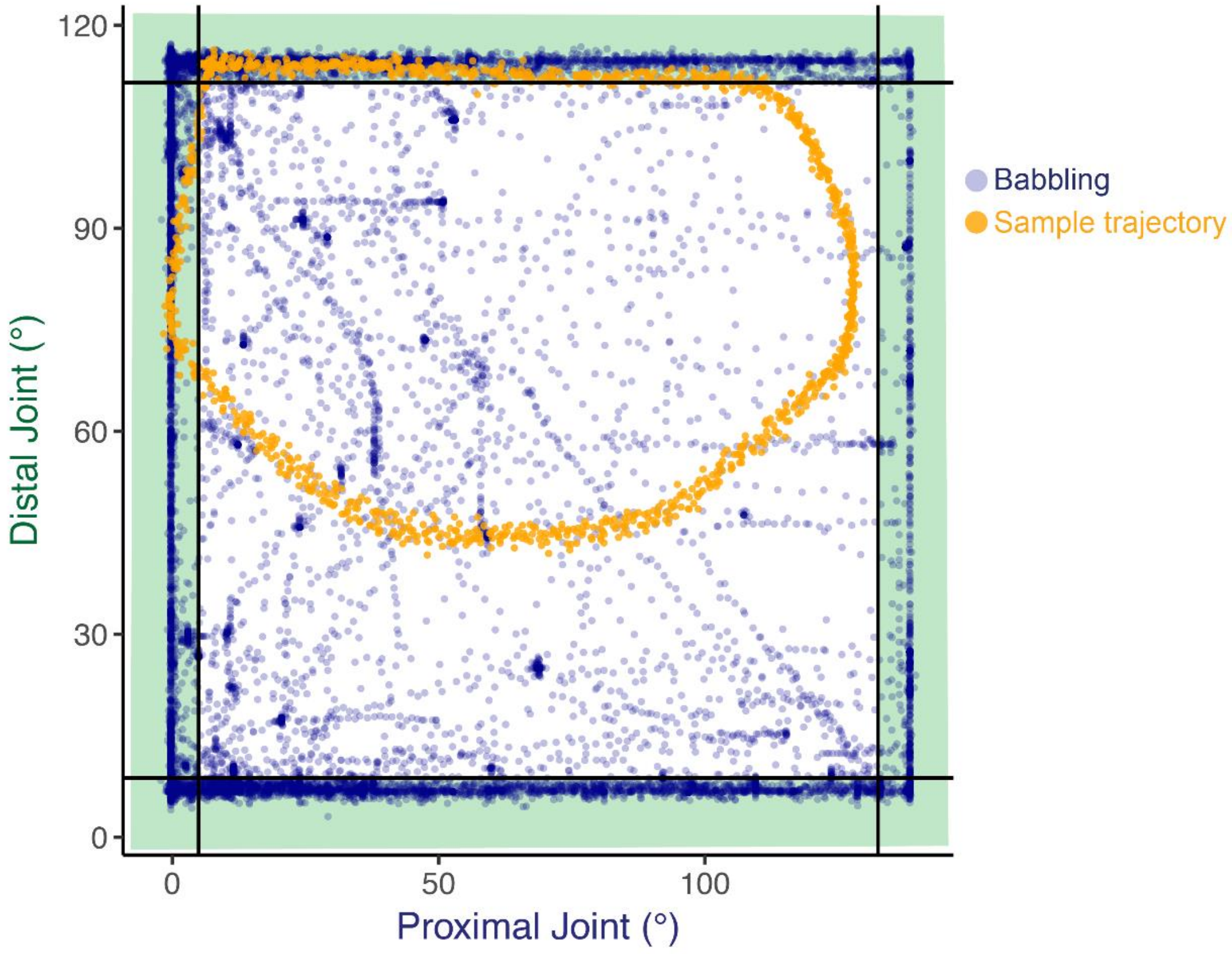

**Figure 6. Joint angle distribution for motor babbling vs. one attempt of a free cyclical movement in air** Motor babbling primarily results in observations at the extremes of the ranges of motion of each joint (23,400 blue points), whereas the desired movement trajectories require exploitation of the relatively unexplored internal region of the joint angle space (1,200 orange points for 16 cycles of a given attempt-excluding the initial transition). The shaded region of this phase plane plot identifies the positions where at least one of the joints was within 5% of its joint limits (black lines).

## Do motor habits improve versatility?

Improvements due to familiarity with a task, however, can be thought of as overfitting that is locally useful, but does not necessarily improve the versatility of the limb in general. We therefore performed a final cross-validation experiment to test whether the improvements of these free cyclical movements in air were solely beneficial to those movements for which the inverse map was trained. Figure 5 shows that an implicit inverse map refined with a set of 30 movements is better at producing different set of movements it has never experienced, than a general implicit inverse map created purely from random motor babbling data. Said differently, having experience with forming a variety of motor habits does not necessarily overfit the implicit map, but rather grants the system deeper knowledge of its own dynamics. That is, familiarity with one's motion capabilities for some tasks seems to inform the execution of other tasks.

**Reward vs. energetic cost**

Our results in Figure 4b are particularly interesting because they show that energy minimization is not an emergent property of this algorithm[59]. If we consider the attempt from each run that performed above threshold as a string of related solutions (e.g., attempts), we see there is drift towards higher performance (by construction), but the family as a whole can be narrow or wide from the perspective of energetic cost. Not surprisingly, there are multiple solutions with similar cost, but nowhere do we see a trend towards energy minimization (i.e., none of the convex hulls are shaped diagonally towards the top right). Conversely, because the solutions for a given run are kinematically similar to each other (they are all increasingly minor modifications of the first limit cycle that crossed the exploration-exploitation threshold), it is interesting to note the different families do not follow some common energetic trend. One could have expected that movements that caused more propulsion would be more energetically costly as they do more work against the treadmill, yet we also do not see such a consistent trend diagonally towards the top left. Thus we are led to conclude that energy consumption is likely most related to how muscles fight each other in the nullspace of the task, as elaborated in detail elsewhere[10,11,60].

**Limitations, opportunities, and future directions**

The G2P algorithm is designed to allow reasonable performance with limited data and no direct feedback, which naturally leads to limitations. The absence of a gradient to follow leads to inconsistent performance where the next attempt may have lower performance, which would be compounded by if the plant has time-varying dynamics.  More crucially, given the non-linear dynamics of even a time-invariant plant, the initial conditions will also affect each attempt. This is why we always begin trials from a same initial configuration. We also exclude the first quarter of each attempt data during the refinements to exclude the transitioning from the initial conditions to the limit-cycle kinematics. Moreover, we do not rely on off-line simulation to explore refinements, but rather always roll out the dynamics in the real physical system. This is contrary to model-predictive control—which would likely produce better performance in certain applications.

However, our fundamental motivation is to understand how biological systems learn to move in a well-enough fashion when they must also execute every attempt using their own bodies. These insights provide a foundation on which we can develop versatile bio-inspired robots. Importantly, for organisms as for machines, there exists a trade-off between improving performance via practice, but where each attempt or exploration carries the risk of, inter alia, a costly injury, fatigue, and wear of tissues (e.g. blisters in the skin, stress fractures, inflammation of tendons, cartilage wear) or parts. In fact, identical repeats of a same task carry the risk of repetitive stress injuries in elite athletes and musicians—and mechanical failure in robots. Variability of reward and execution is therefore a fact of life for organisms, and something G2P uses to its advantage. But even if performance were the main objective of a given robot, G2P can still be very useful by providing a first approximation to a reasonable solution that serves as the critical starting point for subsequent optimization via existing methods.

The use of a limit cycle is a strong assumption in this first use of G2P, which may not be appropriate for every application, but applies well to locomotion, reach, manipulation, swimming and flight. However, we were pleasantly surprised to see that G2P works well even for the longstanding problem of intermittent contact with the ground. The G2P algorithm is unaware that ground contact leads to hybrid and strongly discontinuous dynamics, and we measure no foot forces nor locations relative to ground. Unlike gain scheduling or central pattern generators[61] where control strategies are switched by, or phase-locked to contact, G2P learned to transition across dynamic domains by adjusting its control signals and leveraging the properties of the plant—much as brains likely co-evolved to produce motor actions that leverage the passive viscoelastic properties of musculotendons and skin pads to manage intermittent contacts with the environment and objects, or leverage the physics of inverted pendula for locomotion[62]. We could have mitigated the effects of contact dynamics by tracking contact events or forces. However, we took on the larger challenge of letting G2P autonomously produce a control strategy to handle contacts with the environment using the properties of its actuators and plant.

Data-driven approaches must contend with deciding how much data are enough, or even necessary, for a given level of performance. We did not approach this formally, but rather posed the question at a heuristic at human-scale: Can G2P learn a task after only five minutes of motor babbling, where each motor command changes at an amortized rate of 1 Hz. While G2P succeeded within these constraints, it is natural to ask if more motor babbling would be better or what is the minimal amount of babbling that can yield a functional map. We did not explore optimal stopping rules[63], or apply a Metropolis–Hastings algorithm for a Markov process with ergodic assumptions, parameter distributions, or acceptance probabilities[64]. G2P would certainly benefit from such formal analyses and improvements in the future. In this first application, we set the exploration-exploitation threshold to begin the exploitation phase at 64 mm of treadmill reward (Figures 3b and 4a). The runs in our study crossed the exploration-exploitation threshold after a median of 14 attempts (Figure 4a), with the best reward occurring after 24 attempts (median across replicates). The exploration-exploitation threshold will likely be different for other tasks, as per their dimensionality, nonlinearity, etc.—or even for this same task as we add more tendons and joints. Similarly, a wide set of alteration in system parameters or reinforcement learning algorithm may prove to be efficient when combined with our G2P algorithm. However, for few-shot practical use with real world physics with minimal prior information on the plant, environment, or the task, the G2P algorithm is superior to the state of the art techniques that rely heavily in long term explorations which is only possible with simulations (where the model of the plant and its interactions with the environment is given)[65,66]. This first application of G2P suffices to provide a strong proof-of-principle of its capabilities in using a few-shot approach in producing multiple families of functional controllers for locomotion with a tendon-driven limb. Further biologically realistic refinements, such time delayed sensory feedback during task execution, can only improve its performance.

An important, even critical, takeaway from this work is that—to our knowledge—it is the first to shed light (via both algorithm and physical implementation) on potential biological mechanisms that enable vertebrates to learn to use their bodies within a practical number of attempts to mitigate the risks of injury and overuse—and successfully engage in predator-prey interactions. The ingredients and steps of G2P are all biologically-tenable (i.e., trial-and-error, memory-based pattern recognition, Hebbian learning, experience-based adaptation), and allow us to move away from the reasonable, yet arguably anthropocentric and teleological, concepts dominating computational neuroscience such as cost functions, optimality, gradients, dimensionality reduction, etc.[9–11,54,67]. While those computational concepts are good metaphors, it has been difficult to pin down how one would be able to actually demonstrate their presence and implementation in biological systems[68]. In contrast, G2P can be credibly implementable in biological systems. Our own future direction is to demonstrate its implementation as a neuromorphic neuromechanical system, as we have done for other sensorimotor processes[69,70]. At a conceptual level, this proof of concept of few-shot motor learning has clinical applications. For example, rehabilitation science is increasingly leaning towards mass practice because outcomes tend to be disappointing when the doses of rehabilitation are limited due to cost, personnel, and time[71]. In this context, G2P may provide further insight toward the learning mechanisms that are operating in the nervous system, which can then enable novel clinical strategies.

## Conclusion

G2P produced two important results in the context of the challenging task of controlling a tendon-driven system with sparse data. First and foremost, we demonstrate this biologically- and developmentally-inspired system (in both mechanics and control) uses a few-shot approach to successfully produce free movement and locomotion in spite of data-poor sampling and no real-time feedback, or prior information about the plant or environment. Second, G2P brings novel possibilities to robotics in general as it shows that a few-shot approach to autonomous learning can lead to effective and generalizable control of complex limbs for robotic locomotion and, by extension, manipulation, swimming and flight. Our results demonstrate a powerful biologically-inspired approach that can and should be extended to produce versatile adaptation in spite of changes in the particular target task, mechanics of the robot's anatomy (e.g., damaged limbs), changes in payload, and interaction with the environment (e.g., contact, flow and force fields, etc.). The inherent adaptability of G2P allows extensions to, for example, co-evolve the details of the body and controller to match the requirements of multiple tasks and environments. Ultimately, it can also serve as a conceptual template to empower computational neuroscience studies. This would both advance our understanding of the mechanism that grants vertebrates their enviable versatility and performance, and allow their implementation in a new generation of biologically-inspired robots.

## Materials and Methods

### Physical plant

We designed and built a planar robotic tendon-driven limb with two joints (proximal and distal) driven by three tendons, each actuated by a DC brushless motor. A passive hinged foot allowed natural contact with the ground. We used DC brushless motors as they have low mechanical resistance and are backdrivable. The motor assembly and proximal joint are housed in a carriage that can be lowered or raised to a set elevation for the foot to either reach a treadmill or hang freely in the air (Figure 1). Further considerations and part details can be found in Supplementary Materials.

Tendons

We used three tendons because is the minimal number required for controllability of a planar 2-DOF system (n+1 where n is the number of joints)[10]. During construction, we found that tendons would occasionally go slack and reposition themselves off their tendon channels. Thus, we added PLA conical flanges to the motor collars, redesigned the tendon channels to be deeper. We always applied a baseline tension of 15% of maximal motor activation (similar to muscle tone; Figure 1 a and d) to prevent such problems and because it is necessary for control of tendon-driven systems[39].

Feasible Wrench Set and Design Validation

Joint moment arms and tendon routings were simulated and ultimately built to have adequate endpoint torque and force sets conducive to pressing and walking (Figure 1c). As in[10], a tendon-driven limb's ability to create an output wrench is dependent on the components defining the feasible force space (i.e. the space of all possible output force vectors for a given position):

$w=J^{-T}RF_{max}.a=H.a$

Where $w$ represents the wrench output, $J$ is the Jacobian that maps joint rotations into end-effector directions, $R$ is the moment arm matrix (i.e. the signed leverage each muscle has across each joint), $F_{max}$ is the maximal tension possible of each tendon and $a$ is a unit vector representing the by-tendon fraction of total activation. This equation can be simplified to $Ha$. By systematically evaluating all binary combinations for the elements in $a$, the resultant wrenches give rise to a feasible force set. It is important to preserve the physical capability of the tendon routing through the many iterations of limb design, so at each design phase we computed these sets for different positions throughout the limb propulsive stroke. Downward force ($-f_y$) served as our primary metric (which is needed to eventually support weight when this leg design is used on a quadruped). Linear programming and computational geometry[10] corroborated that the limb was provisioned with motors capable of generating downforce bigger than the system's weight, with ample affordance so it could propel the body forward[72].

Mechanical considerations

The carriage was attached to a wooden support structure, via linear-bearing and slide rails to adjust its vertical position. A clamp prevented sliding once the vertical position was set. Sandpaper was glued to the footpad and in strips across the treadmill to improve traction (Figure 1 a and b).

Data acquisition

The control system had to provide research-grade accuracy and consistent sampling to enable an effective hardware test of G2P. A Raspberry Pi (Raspberry Pi Foundation, Cambridge, U.K.) served as a dedicated control loop operator—issuing commands to the motors, sensing angles at each of the proximal and distal joints, and recording the treadmill band displacement (Figure 1 a and b). Furthermore, the electrical power consumption for each motor was measured at 500Hz using current-sensing resistors in parallel with the motor drivers, calculating the Watt-hours over each inter-sample-interval, and reporting the amortized mean watts for the entire attempt. All commands were sent, and data received, via wireless WiFi communication with the Raspberry Pi as csv files.

Running the system

The limb is placed in a consistent starting posture before activations are run to minimize variance in the initial conditions of the plant. To aid development, a live-streaming video feed was designed for real-time visualization on any computer on the network (See Movie S1). A computer sends a control sequence to the Raspberry Pi, and after it is successfully run, the computer receives (i) the paired input-to-output data in csv format for iterative analysis or training, (ii) the net distance (mm) covered over the course of the entire action, and (iii) the amortized power the system consumed during the trial. Once data are collected, samples are interpolated using their corresponding time labels to combat the nonuniform inter-sample interval of 78±5Hz. Prescribed activation trajectories are also served at this rate. The pipeline for data acquisition was designed with Python 3.6.

## Simulations

We first prototyped our methods in simulation using a double pendulum model of a tendon-driven limb. Similar to the physical system, our method proved to be efficient in the simulation and yielded comparable results (Figures S2 and S3). These simulations were kept isolated from the physical implementation, and its results were never used as seeds for the physical implementation.

## Learning and control algorithm

Learning and control in this first implementation of the G2P algorithm happens at two levels: (i) inverse mapping and refinement (the lower-level control) and (ii) the reward-based reinforcement learning algorithm (the higher-level control). The lower-level is responsible for creating an inverse map that converts kinematics into viable control sequences (motor commands). The higher-level control is responsible for reward-driven exploration (reinforcement learning) of the kinematics which are further passed to the lower-level control and ultimately run through the system.

### Inverse mapping and refinements

The lower-level control relies on two phases. As system is provided with no prior information on its dynamics, topology, or structure, it will first explore it dynamics in a general sense by running random control sequences to the motors, which we call motor babbling. After 5 minutes of motor babbling, the system creates the initial inverse map using data collected from particular task-specific explorations, which we refer to as task-specific adaptation. This transition from motor babbling to adaptation to a particular task is the reason we refer to this algorithm as General to Particular or G2P.

## Motor Babbling

During this phase, the system tries random control sequences and collects the resulting limb kinematics. A Multi-Layer Perceptron (MLP) Artificial Neural Network (ANN) is trained with this input-output set to generate an inverse map between the system inputs (here, motor activation levels) and desired system outputs (here, system kinematics: joint angles, angular velocities, and angular accelerations). Although sparse, data from these inputs and outputs suffice for the ANN to create an approximate general map based on the system's dynamics.

### Random activation values for the babbling

The motor activation values (control sequences) for motor babbling were generated using two pseudo-random number generators (uniformly distributed). The first random number generator provides a 1/fs time probability for the activation level to move from one command level to another. The second number defined the activation value of the next state with sampling from a range of 15% to 100% activation (see Tendons subsection). The resulting command signals were stair-step transitions in activations to each motor.

### Structure of the Artificial Neural Network

The ANN representing the inverse map from 6-dimesional limb kinematics to 3-dimensional motor control sequences has 3 layers (input, hidden, and output layers) with 6, 15, and 3 nodes, respectively. The transfer functions for all nodes were selected as the hyperbolic tangent sigmoid function (with a scaling for the output layer to keep it in the range of the outputs). ANN mapping was performed in MATLAB (Neural Network ToolBox; MathWorks, Inc., Natick, MA).

Task based refinements

Motor babbling yields sample observations distributed across a wide range of dynamics, but still represents a sparse sampling of the range of state-dependent dynamical responses of the double pendulum (Figure 6). As a result, this initial inverse map ($ANN_0$, Figure 3) can be further refined when provided with more task-specific data.

The higher-level control will initiate the exploration phase using $ANN_0$. However, with each exploration, the system is exposed to new, task-specific data, which is thereby appended to the database and incorporated into the refined $ANN_K$ maps (Figure 3). This refinement is achieved by using the current weights as the initial weight of the refined ANN and training it on the cumulative data after each attempt. It is important to note that refinements can update the map's validity only to a point; if major changes to the plant are experienced (changing the tendon routings or the structure of the system) the network would likely need to re-train on new babbling data. However, we found that motor babbling done strictly while the limb was suspended in air nevertheless worked well when it was used to produce intermittent contact with the treadmill to produce locomotion on the treadmill.

The reinforcement learning algorithm

A two-phase reinforcement learning approach is used to systematically explore candidate system dynamics, using a 10-dimensional limit cycle feature vector, and converge to the ones with highest reward. Before explaining these phases, we detail how the candidate kinematics are generated by a limit cycle parametrized using 10 free parameters.

Limiting the search space and creating limit-cycle feature vectors

At each step of the reinforcement algorithm, the policy must produce a candidate set of kinematics. A locomotor task is a 1-DOF limit cycle embedded in the 6-dimensional space of two joint angles, angular velocities and angular accelerations. Beginning with a circle centered on the origin of the angle-angle space, we defined ten equally-distributed spokes (Figure 2c). We can then set the lengths of each spoke (i.e., the 10 free parameters) to define an arbitrary closed path that defines the time history of angle changes, which remains a smooth and differentiable limit cycle. These ten lengths of the spokes are the 10-dimensional limit cycle feature vector. Assuming movement between these ten points with equal inter-point duration then defines the associated angular velocities and accelerations, which fully describe a cyclical limb movement. This 6-dimensonal target limb movement can be mapped into the associated control sequences to produce it by the inverse map. Those control sequences are concatenated 20 times and fed to the motors to produce 20 back-to-back repetitions of the cyclical movement. These features were bounded in [0.15-1] for the treadmill task and [0.2-0.8] during the free cyclical movements experiments (to provide more focused task specific trajectories).

Exploration phase

Exploring random attempts across the 10-dimensional feature vector space (uniform at random in [0.15-1]) eventually will produce solutions which yield a treadmill reward. Exploration continues until either the reward is higher than a predefined threshold or stopped when a maximal run number is surpassed (a failure).

Exploitation phase

Once the reward passes the threshold, the system will select a new feature vector in the vicinity of the feature vector from a 10-dimensional Gaussian distribution, with each dimension centered at the threshold-jumping solution. Much like a Markov process, with each successful attempt, the 10-dimensional distribution will be centered on the values of the feature vector which yielded the best reward thus far. The standard deviation of these Gaussian distributions is inversely related to the reward (the distribution will shrink as the system is getting more reward). The minimal standard deviation is bounded at 0.03.

Between every attempt, the ANN's weights are refined with the accumulated dataset (from motor babbling and task-specific trajectories) regardless of the reward or reinforcement phase. This reflects the goal for our system to learn from every experience.

## Acknowledgments

We thank H. Zhao for his support in designing and manufacturing the physical system as well as support in the analysis of the limb kinematics, S. Kamalakkannan for support in designing and implementing the data acquisition system, and Y. Kahsai for the illustration of Figures 2 and 3.

**Funding:** Research reported in this publication was supported by the National Institute of Arthritis and Musculoskeletal and Skin Diseases of the National Institutes of Health under Awards Number R01 AR-050520 and R01 AR-052345 to F.J.V.-C. This work was also supported by the Department of Defense CDMRP Grant MR150091 to F.J.V.-C. We acknowledge additional support for A.M for Provost and Research Enhancement Fellowships from the Graduate School of the University of Southern California and fellowships for D.U.-M. from the Consejo Nacional de Ciencia y Tecnología (Mexico), and for B.C. from the NSF Graduate Research Fellowship Program. The content of this endeavor is solely the responsibility of the authors and does not necessarily represent the official views of the National Institutes of Health, the Department of Defense, The National Science Foundation nor the Consejo Nacional de Ciencia y Tecnología.

**Author contributions:** All authors contributed to the conception and design of the work, and writing of the manuscript. A.M. also led the development of the G2P algorithm, D.U.-M. also led the construction of the robotic limb, B.C. also led the data acquisition and analysis. F.V-C. also provided general direction for the project. All authors have approved the final version of the manuscript and agree to be accountable for all aspects of the work. All persons designated as authors qualify for authorship, and all those who qualify for authorship are listed.

**Competing interests:** The authors have no competing interests.

**Data and materials availability:** The code for our algorithm and the STL files for 3D printing the physical system are available at a GitHub repository (to be provided on the typesetter's version).